%% file: main.tex
\documentclass[nohyperref]{article}
\pdfoutput=1

\usepackage[accepted]{icml2022}

\usepackage{amsmath,amsfonts,amssymb}
\usepackage{mathtools}
\usepackage{microtype}
\usepackage{booktabs}
\usepackage{graphicx}
\usepackage{hyperref}
\usepackage{url}

\usepackage[font=small,skip=2ex,labelfont=it]{caption}
\usepackage{subcaption}
\usepackage{paralist}

\usepackage{multirow}
\usepackage{longtable}
\usepackage{makecell}

\usepackage{etoolbox}

\newrobustcmd{\B}{\bfseries}

\icmltitlerunning{Translatotron 2: High-quality direct speech-to-speech translation with voice preservation}

\begin{document}

\twocolumn[
\icmltitle{Translatotron 2: High-quality direct speech-to-speech translation\\ with voice preservation}

\icmlsetsymbol{equal}{*}

\begin{icmlauthorlist}
\icmlauthor{Ye Jia}{gg}
\icmlauthor{Michelle Tadmor Ramanovich}{gg}
\icmlauthor{Tal Remez}{gg}
\icmlauthor{Roi Pomerantz}{gg}
\end{icmlauthorlist}

\icmlaffiliation{gg}{Google Research}

\icmlcorrespondingauthor{Ye Jia}{jiaye@google.com}

\icmlkeywords{Speech-to-speech translation}

\vskip 0.3in
]

\printAffiliationsAndNotice{}  %

\input{0-abstract}

\input{1-introduction}
\input{2-method}

\input{3-experiments}

\input{4-conclusion}

\bibliography{references}
\bibliographystyle{icml2022}

\newpage
\input{9-appendix}

\end{document}

%% file: 0-abstract.tex
\begin{abstract}
We present \emph{Translatotron 2}, a neural direct speech-to-speech translation model that can be trained end-to-end. Translatotron 2 consists of a speech encoder, a linguistic decoder, an acoustic synthesizer, and a single attention module that connects them together. Experimental results on three datasets consistently show that Translatotron~2 outperforms the original Translatotron by a large margin on both translation quality (up to $+$15.5 BLEU) and speech generation quality, and approaches the same of cascade systems.
In addition, we propose a simple method for preserving speakers' voices from the source speech to the translation speech in a different language. Unlike existing approaches, the proposed method is able to preserve each speaker's voice on speaker turns without requiring for speaker segmentation. Furthermore, compared to existing approaches, it better preserves speaker's privacy and mitigates potential misuse of voice cloning for creating spoofing audio artifacts.
\end{abstract}

%% file: 1-introduction.tex
\section{Introduction}

Speech-to-speech translation (S2ST) is highly beneficial for breaking down communication barriers between people not sharing a common language. Conventional automatic S2ST systems are composed of a cascade of three components: automatic speech recognition (ASR), text-to-text machine translation (MT), and text-to-speech (TTS) synthesis \citep{lavie1997janus, wahlster2000verbmobil, nakamura2006atr}.
In the past few years, direct speech-to-text translation (ST) is rapidly emerging, and has outperformed the cascade of ASR and MT \citep{weiss2017sequence,jia2019leveraging,di2019one,mccarthy2020skinaugment,ansari2020findings,wang2021large,anastasopoulos2021findings}, which makes the cascade of ST and TTS as S2ST feasible \citep{jia2019direct}. %

Recently, works on S2ST without relying on intermediate text representation are emerging, such as end-to-end direct S2ST \citep{jia2019direct,kano2021transformer} and cascade S2ST based on discrete speech representation \citep{tjandra2019speech,zhang2020uwspeech,lee2021direct,lee2021textless,ma2021direct}. 
Compared to text-centric cascaded systems, although such approaches require parallel S2ST data, which is scarce, they have the potential advantages of:
\begin{inparaenum}[1)]
    \item Preserving paralinguistic and non-linguistic information during translation, such as speaker's voice \citep{jia2019direct}, emotion and prosody;
    \item Supporting languages without written form, or being able to be trained without transcription of speech \citep{tjandra2019speech,zhang2020uwspeech,lee2021direct,lee2021textless};
    \item Reduced computational requirements and lower inference latency \citep{lee2021direct};
    \ifdefined\isaccepted 
    \item Avoiding error compounding across sub-systems \citep{jia2022cvss};
    \else
    \item Avoiding error compounding across sub-systems;
    \fi
    \item Easier on handling contents that do not need to be translated, such as names and proper nouns \citep{jia2019direct}.
\end{inparaenum}

Among these works, Translatotron \citep{jia2019direct} is the first model that is able to directly translate speech in one language to speech in another language. It obtained reasonable translation quality and high naturalness in the predicted translation speech, and is able to preserve speakers' voices during the speech translation.
However, the translation quality from Translatotron still underperforms cascade baselines by a large margin, and the translation speech it produces suffers from over-generation issues, such as babbling and long pause. %
Such weaknesses make this model not yet practical for production.
Nevertheless, it remains the state-of-the-art of end-to-end direct S2ST.

In this paper, we first tackle the performance gap between end-to-end direct S2ST and cascade S2ST.
We propose \emph{Translatotron 2},
a novel direct S2ST model that is able to be trained end-to-end. We conduct experiments on three S2ST datasets, including multilingual S2ST. The results consistently suggest that Translatotron~2 significantly outperforms Translatotron in terms of both translation quality (up to $+$15.5 BLEU) and speech generation quality, and approaches the same of cascade S2ST.
When a simple data augmentation \emph{ConcatAug} is used, the translation quality gap on the Fisher Spanish-English corpus \citep{post2013improved} is reduced from 16.4 to 0.4 BLEU.
These results are the first time that end-to-end direct S2ST approaches cascade S2ST.

In addition, we propose a simple method for preserving speakers' voices during S2ST without relying on any speaker representation (ID or embedding). The proposed method enables Translatotron 2 to preserve each speaker's voice on speaker turns without requiring for speaker separation, which is the first of its kind.
Furthermore, compared to existing approaches of voice preservation, the proposed method better preserves speaker's privacy \citep{pathak2012privacy} and mitigates potential misuse of voice cloning for creating spoofing audio artifacts.

Audio samples from Translatotron 2 are
\ifdefined\isaccepted 
available online.\footnote{\url{https://google-research.github.io/lingvo-lab/translatotron2/}}
\else
available.\footnote{In the supplementary materials.}
\fi

\section{Related works}

\paragraph{S2ST}

Until very recently, automatic S2ST systems are typically composed of a cascade of ASR, MT, and TTS components \citep{lavie1997janus, wahlster2000verbmobil, nakamura2006atr,itu-f745}.
Translatotron \citep{jia2019direct} is the first direct S2ST model, which is a sequence-to-sequence model trained end-to-end in a multi-objective task. It has shown reasonable translation quality and speech naturalness, but still underperformed a baseline of ST $\to$ TTS cascade by a large margin. It also demonstrated the capacity of preserving speakers' voices during the translation, by leveraging a speaker encoder separately trained in a speaker verification task \citep{wan2018generalized,jia2018transfer}.

A few recent works proposed cascade S2ST systems using learned discrete speech representation as the intermediate representation instead of text or phoneme.
\citet{tjandra2019speech} introduced such an S2ST system that first translated the source speech into a discrete representation of the target speech which was predicted from a separately trained VQ-VAE \citep{oord2017neural}, then constructed the target speech spectrogram from the discrete representation using the VQ-VAE decoder.
\citet{zhang2020uwspeech}
additionally trained the VQ-VAE jointly with a supervised phoneme recognition objective in different languages.
\citet{lee2021direct,lee2021textless} used a separately trained vocoder to directly predict waveform from the discrete representation without relying on spectrogram; for the best performance, this vocoder included a duration predictor, akin to a generative TTS model.
All these works require multiple components being trained in multiple steps, but are not able to be trained end-to-end. %
Another potential limitation of such an approach is that it may not be effective in preserving paralinguistic and non-linguistic information. Oppositely, it can be desired that such variation be removed in the discrete representation \citep{lee2021textless}.

\citet{kano2021transformer} introduced an end-to-end S2ST model with a cascade of three autoregressive decoders, and used pre-trained MT and TTS models as teacher models to facilitate the training of the end-to-end model. It requires pre-trained ASR, MT, and TTS models, and the end-to-end model itself has to be trained in multiple steps.

While most of these works conducted experiments using synthetic datasets with translation speech in a clean single speaker's voice, \citet{jia2019direct,lee2021textless} conducted experiments using %
multi-speaker human recordings.

Although these recent works generated speech translation in novel ways without relying on TTS subsystems, only a few of them \citep{jia2019direct,lee2021direct} have evaluated the perceptual quality (e.g. naturalness) of the produced speech translation, which is critical to S2ST \citep{wagner2019speech,salesky2021assessing}, with the rest focused only on the translation quality.

\paragraph{TTS}
Translatotron uses a decoder similar to the Tacotron~2 TTS model \citep{shen2018natural}, which is an attention-based autoregressive decoder. Due to the flexibility of the attention mechanism, they both suffer from robustness issues such as over-generation. Recent TTS models such as FastSpeech \citep{ren2019fastspeech,ren2020fastspeech2} and Non-Attentive Tacotron (NAT) \citep{shen2020non} demonstrated that replacing the attention module with a duration-based upsampler yields
\ifdefined\isaccepted 
more robust synthesized speech, as quantitatively evaluated at a large scale in \citet{shen2020non}.
\else
more robust synthesized speech.
\fi
The synthesizer component in this work resembles these works.

\paragraph{Voice conversion and anti-spoofing}

The performance of voice conversion has progressed rapidly in the recent years, and is reaching a quality that is hard for automatic speaker verification (ASV) systems to detect \citep{yi2020voice}. ASVspoof 2019 \citep{todisco2019asvspoof, wang2020asvspoof} found that it was challenging to detect spoof audios generated from a zero-shot voice cloning TTS model \citep{jia2018transfer}, which was followed by the original Translatotron for preserving speakers' voices during S2ST.  
Such progress poses concerns on related techniques being misused for
creating spoofing artifacts. %
We propose a new  voice preservation method for S2ST with the motivation of avoiding such potential misuse.

%% file: 2-method.tex
\section{Translatotron 2}
\label{sec:method}

\begin{figure*}[t]
    \centering
    \begin{subfigure}[b]{0.4\textwidth}
        \hfill%
        \includegraphics[height=2.6in]{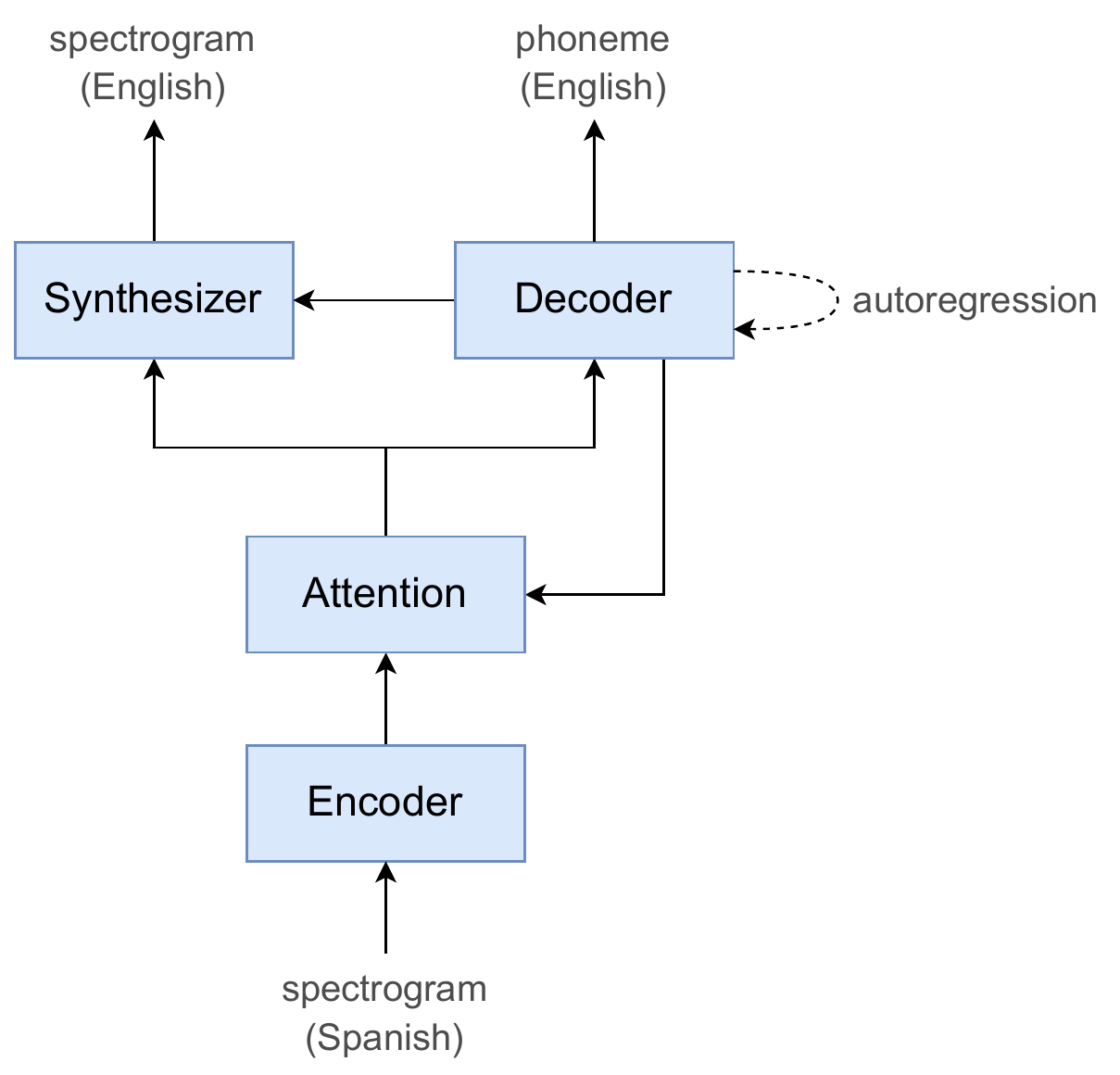}
        \caption{Overview of Translatotron 2.}
        \label{fig:model-overall}
    \end{subfigure}%
    \begin{subfigure}[b]{0.4\textwidth}
        \centering
        \includegraphics[height=2.4in]{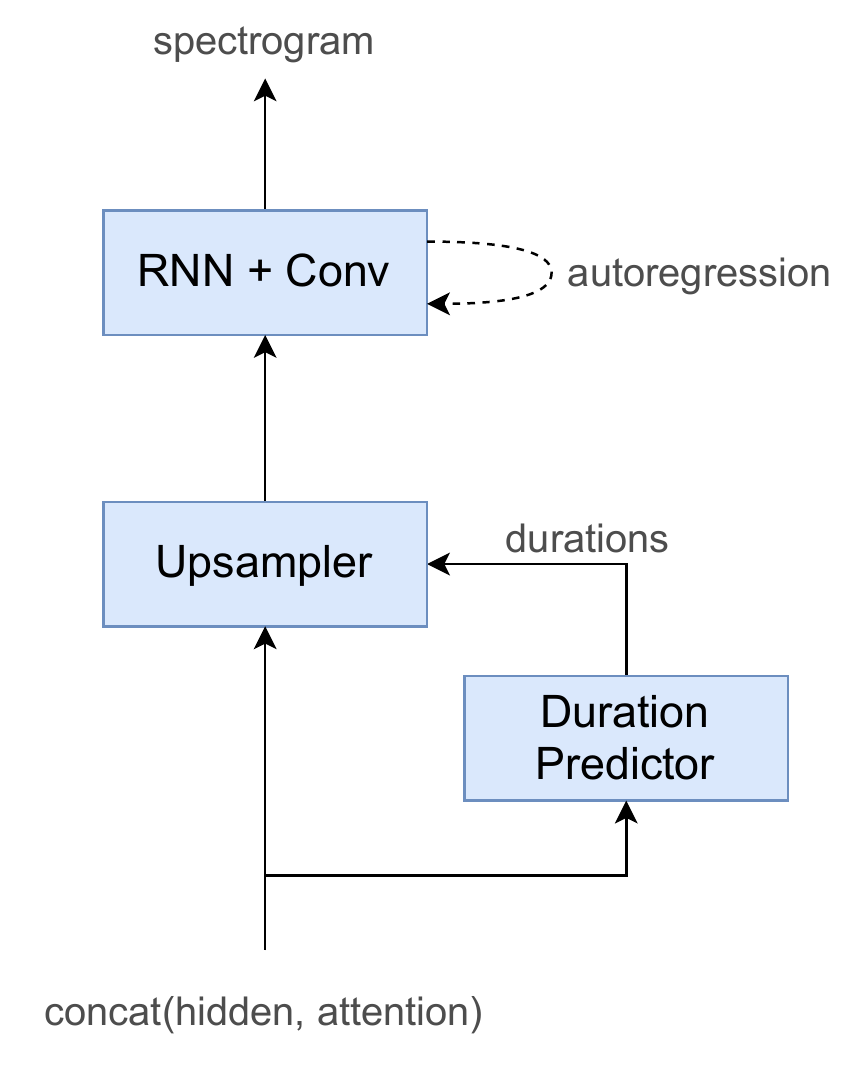}
        \caption{Synthesizer of Translatotron 2.}
        \label{fig:model-synthesizer}
    \end{subfigure}
    \caption{A Translatotron 2 model that translates Spanish speech into English speech.}
    \label{fig:model}
\end{figure*}

We designed the architecture of Translatotron~2 to address three performance bottlenecks existing in the original Translatotron: 1) The utilization of the auxiliary textual supervision during training is suboptimal, namely, the attention alignment learned by the auxiliary ST task does not directly contribute to the main S2ST task; 2) The challenge posed by modeling the translation alignment between two very long spectrogram sequences using the attention mechanism; 
3) Attention-based speech generation is known to suffer from robustness issues such as over-generation and under-generation \citep{shen2020non, ren2019fastspeech, he2019robust, zheng2019forward, battenberg2020location}.

We addressed these bottlenecks by designing a novel S2ST model architecture composed of a speech encoder, a linguistic decoder, an acoustic synthesizer, and a single attention module connecting them together (Figure~\ref{fig:model-overall}). 
The model is jointly trained with a speech-to-speech translation objective and a speech-to-phoneme translation objective.

The following subsections describe each component of Translatotron 2. Note that as shown in the ablation studies in Sec.~\ref{s:ablation}, while the specific architectural choices of these components help the performance of Translatotron~2, the primary improvement comes from the high-level architecture rather than the choice of each individual component.

\subsection{Speech encoder}

The encoder of Translatotron 2 takes the mel-spectrogram of the source speech as the input, and produces a hidden representation which encodes both linguistic  and acoustic information from the source speech. We use Conformer \citep{gulati2020conformer} as the architecture of the encoder. It first subsamples the input mel-spectrogram with a convolutional layer, and then processes it with a stack of Conformer blocks. Each Conformer block is composed of a feed-forward layer, a self-attention layer, a convolution layer, and a second feed-forward layer.
SpecAugment \citep{park2019specaugment} is applied at the training time as data augmentation.

\subsection{Linguistic decoder}

The autoregressive decoder is responsible for producing linguistic information in the translation speech. 
It takes the encoder output through the attention module, and predicts a phoneme sequence corresponding to the translation speech. We use an LSTM stack \citep{hochreiter1997long} as the decoder, assisted with regularization including Zoneout \citep{krueger2017zoneout} and label smoothing \citep{szegedy2016rethinking}.
The combination of the encoder, the decoder, and the attention module is similar to a typical ST model, except that it predicts phonemes instead of subword tokens.

\subsection{Acoustic synthesizer}

The synthesizer is responsible for acoustic generation of the translation speech. It takes the
intermediate
 output from the decoder (before a final projection and softmax for phoneme prediction), 
as well as the context output from the attention as its input, and generates a mel-spectrogram corresponding to the translation speech. It is similar to the decoders in typical neural TTS models.
The predicted mel-spectrogram can be converted to waveform using an estimation algorithm such as \citet{griffin1984signal} or a neural vocoder such as WaveRNN \citep{kalchbrenner2018efficient}.

We use the duration-based autoregressive synthesizer from the NAT \citep{shen2020non} TTS model (Figure~\ref{fig:model-synthesizer}). It first predicts durations for each elements in the input sequence, then temporally upsamples the input sequence based on the predicted durations. After that, an LSTM stack is used for generating the target spectrogram without altering the sequence length. A final residual convolutional block further refines the generated spectrogram. Unlike in NAT, we do not supervise the duration prediction on per-phoneme duration labels, to 
avoid additional requirement on the training data.
Instead, an $L^2$ loss on the total predicted duration of the entire utterance is used (similar to the ``na\"ive approach'' of unsupervised duration modeling in \citet{shen2020non}).

\subsection{A single attention}
\label{sec:single-attention}

It is critical that Translatotron~2 utilizes a single attention module for both the linguistic decoder and the acoustic synthesizer. This attention models both linguistic and acoustic alignments between the source and the target speeches. A multi-head attention \citep{vaswani2017attention} is used.

The queries to this attention are from the linguistic decoder. As a result, unlike in the original Translatotron, this attention does not directly model the translation alignment between two very long spectrogram sequences. Instead, it models the alignment between a source spectrogram sequence and a shorter target phoneme sequence, which is significantly easier to learn.

In the meantime, the attention provides acoustic information from the source speech to the synthesizer, summarized at per-phoneme level. Such summarized acoustic information is not only usually sufficient for speech generation but also eases the duration prediction per-phoneme because it is of the same granularity.
Because a single attention is used, the linguistic and acoustic information seen by the synthesizer is synchronized temporally. Such synchronization enables Translatotron~2 to preserve paralinguistic and non-linguistic information at fine granularity, such as preserving each speaker's voice on speaker turns (Sec. \ref{sec:speaker-turns}).

Although the synthesizer takes attention output as part of its input, the attention is not driven (i.e. queried) by the synthesizer. As a result, while it benefits from the attention on obtaining aligned acoustic information from the source speech, it does not suffer from robustness issues as in typical attention-based speech synthesis models.

\begin{table*}[t]
\centering
\begin{small}
\caption{Datasets for experiments with translation speech in a single-speaker's voice.}
\begin{tabular}{lccc}
    \toprule
     & \makecell{Conversational\\ \citep{jia2019leveraging}} & \makecell{Fisher Es-En\\ \citep{post2013improved}} & \makecell{CoVoST 2\\ \citep{wang2020covost}} \\
    \midrule
    Languages  & es$\to$en & es$\to$en & es, fr, de, ca $\to$ en \\
    Domain & Read, short-form & Telephone conversation & Read, short-form \\
    Source sample rate  & 16-48 kHz & 8 kHz & 48 kHz \\
    Utterance pairs & 979k & 120k& 321k \\
    Source hours & 1,400 & 127 & 476 \\
    Target hours & 619 & 96 & 296 \\
    Target synthesized by  & Tacotron 2 + Griffin-Lim & Parallel WaveNet  & PnG NAT  + WaveRNN \\
    \bottomrule
\end{tabular}
\label{tbl:datasets}
\end{small}
\end{table*}

\section{Voice preserving}
\label{sec:voice-retention}

\begin{figure}[t]
  \centering
  \includegraphics[width=1.0\linewidth]{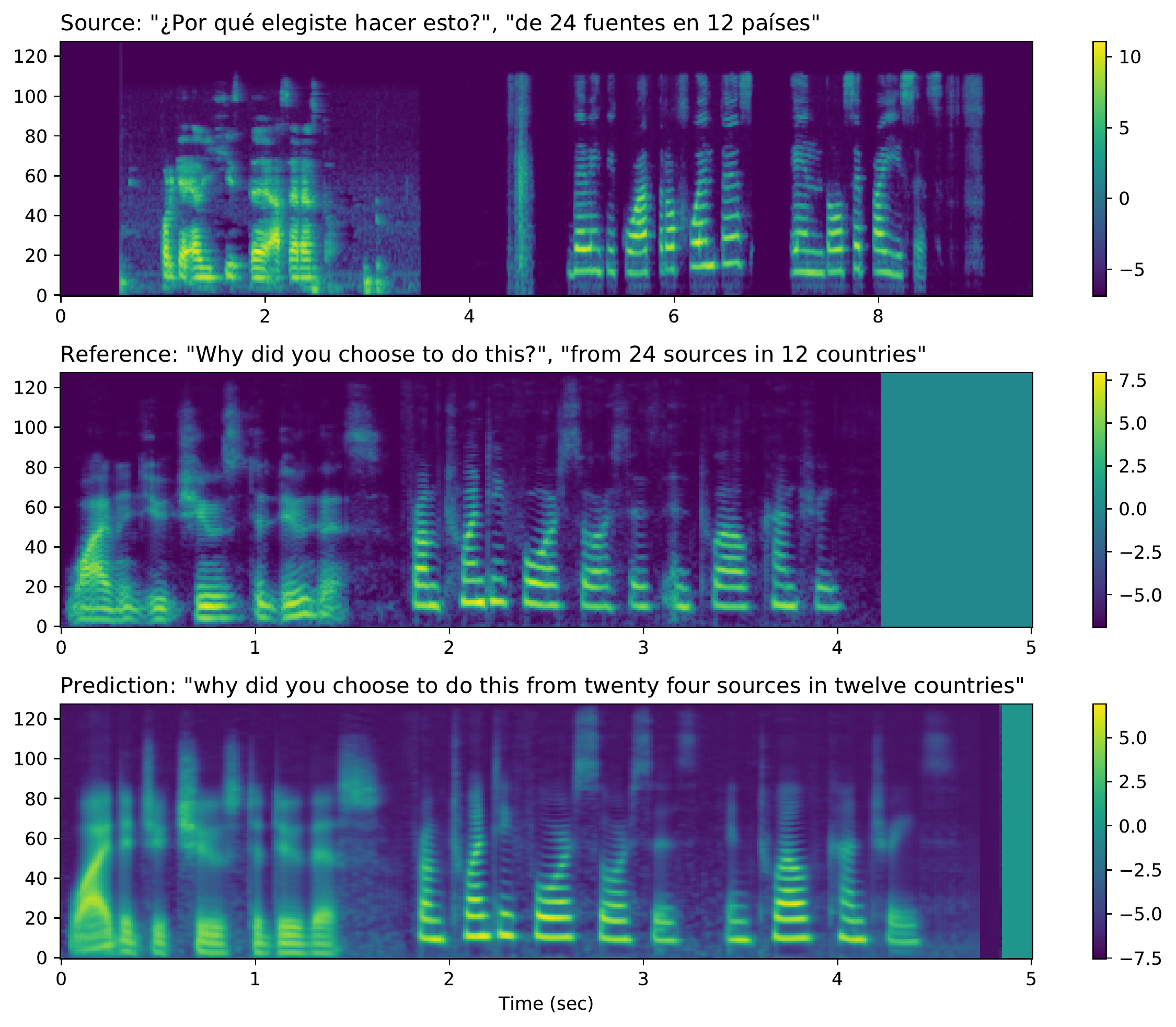}
  \caption{Sample mel-spectrograms on input with speaker turns. The input speech includes an utterance from a male speaker followed by another utterance from a female speaker. Translatotron~2 preserves the voices of each speaker in the translation speech.%
  }
  \label{fig:speaker-switching}
\end{figure}

The original Translatotron \citep{jia2019direct} demonstrated the capacity of preserving source speakers' voices in the translation speech, 
by conditioning its synthesizer on a speaker embedding generated from a separately trained speaker encoder.
In fact, it is capable of generating the translation speech in a different speaker's voice, as long as a clip of the target speaker's recording is used as the reference audio to the speaker encoder, or the embedding of the target speaker is directly available. While this is impressively powerful, it can potentially be misused for generating spoofing audio with arbitrary content, posing a concern for production deployment. %

To mitigate such risks, we propose a new approach for preserving speaker's voice during S2ST, so that the trained models are restricted to preserving the source speaker's voice, but not able to generate speech in a different speaker's voice. In addition, this approach enables S2ST models to preserve each speaker's voice on input speech with speaker turns, without requiring for speaker segmentation.

\subsection{Training-time voice transferring}
\label{sec:multispeaker_data}

In our approach, the key to restrict S2ST models to voice preservation but not arbitrary voice cloning (from a different speaker) is to move the powerful voice transferring to only happen at the training time (or the training data preparation time) but not the inference time. In contrast, it happens at both the training time and the inference time in the original Translatotron. 

To preserve speakers' voices across translation, we train S2ST models on parallel utterances with the same speaker's voice on both sides. Such a dataset with human recordings on both sides is extremely difficult to collect, because it requires a large number of fluent bilingual speakers. Instead, we use a TTS model capable of cross-lingual voice cloning to synthesize such training targets.

We modified the PnG NAT \citep{jia2021png,shen2020non} TTS model by incorporating a separately trained speaker encoder \citep{wan2018generalized} in the same way as in \citet{jia2018transfer}, and trained it on the LibriTTS corpus \citep{zen2019libritts}. The resulting TTS model is capable of zero-shot voice transferring, but synthesizes in a better quality and more robust than \citet{jia2018transfer}.\footnote{A detailed description of this zero-shot voice transferring TTS model is available in our follow-up work \citep{jia2022cvss}.}
We used this model to synthesize translation speech in the source speaker's voice as the training targets in our experiments.
Other TTS models capable of cross-lingual voice modeling, such as \citet{zhang2019learning,chen2019cross,xin2021disentangled}, could also be utilized.

\subsection{Speaker turns}
\label{sec:speaker-turns}

Because the single attention module provides
linguistic and acoustic information temporally synchronized (Sec.~\ref{sec:single-attention}), Translatotron~2 is theoretically capable of voice preservation in complicated scenarios such as speaker turns.
However, proper training data with speaker turns is required to demonstrate such capacity, which is difficult to obtain. We propose a simple data augmentation to enable such training.

\paragraph{ConcatAug}
To enable direct S2ST models to preserve each speaker's voice for input with speaker turns, we augmented the training data by randomly sampling pairs of training examples and concatenating the source speech, the target speech, and the target phoneme sequences to construct new training examples. The resulting new examples contain two speakers' voices in both the source and the target speech, which enables the model to learn on examples with speaker turns. See Figure~\ref{fig:speaker-switching} for an example of such concatenation and the prediction from Translatotron~2 on it.

Such augmentation does not only enable the model to learn voice retention on speaker turns, but also increases the diversity of the speech content as well as the complexity of the acoustic conditions in the training examples, which may further improve the translation quality of the model, especially on small datasets (Sec.~\ref{sec:exp-translation-quality}).
\citet{narayanan2019recognizing} uses a similar augmentation but in a more complicated fashion, for improving ASR performance on multi-speaker inputs.

%% file: 3-experiments.tex
\section{Experiments}

We conducted experiments on three datasets, including two Spanish$\to$English datasets and a multilingual$\to$English dataset. All datasets use TTS synthesized target speech in 24 kHz sample rate. The phonemes used at training time were converted from the transcripts using a proprietary G2P system. See Table~\ref{tbl:datasets} for the details of each dataset.
We evaluated the translation quality, naturalness and robustness of the produced translation speech, as well as speaker similarity for voice preservation.
\ifdefined\isaccepted 
All models were implemented using the Lingvo framework \citep{shen2019lingvo}.
\fi
A comprehensive table of hyper-parameters is available in Appendix~\ref{a:hparams}.

\subsection{Translation quality}
\label{sec:exp-translation-quality}

\begin{table*}[t]
\centering
\begin{small}
\caption{Performance of S2ST in a single speaker's voice. BLEU were computed with 1 reference for the Conversational test set, and with 4 references for the Fisher test set.}
\begin{tabular}{l@{\hspace{0.1em}}rc@{\hspace{0.1em}}rrc@{\hspace{0.1em}}r}
    \toprule
    & \multicolumn{3}{c}{Conversational} & \multicolumn{3}{c}{Fisher Es-En} \\
    \cmidrule(r){2-4}\cmidrule(l){5-7}
    & BLEU & MOS & UDR (\%) & BLEU & MOS & UDR (\%) \\
    \midrule
    \kern-0.5em\emph{End-to-end direct S2ST:} \\
    Translatotron 2                               & \B 55.6 & \B 4.21 $\pm$ 0.06 & 0.16 & 42.4 & \B 3.98 $\pm$ 0.08 & \B 0.07 \\
    \quad + ConcatAug                             & 55.1 & 4.19 $\pm$ 0.06 & \B 0.13 & \B 42.9 & 3.79 $\pm$ 0.09 & 0.14 \\
    Translatotron                                 & 50.4 & 4.15 $\pm$ 0.07 & 0.69 & 26.9 & 3.70 $\pm$ 0.08 & 0.48 \\
    \midrule
    Cascade (ST $\to$ TTS)                        & 58.8 & 4.31 $\pm$ 0.06 & 0.21 & 43.3 & 4.04 $\pm$ 0.08 & 0.13 \\
    Reference (synthetic)                               & 81.9 & 3.37 $\pm$ 0.09 & 0.43 & 88.6 & 3.95 $\pm$ 0.07 & 0.07 \\
    \midrule
    \multicolumn{4}{l}{\kern-0.5em\emph{Discrete representation-based cascade S2ST:}} \\
    \citet{zhang2020uwspeech} (trained w/o text)           &    - &               - & - &  9.4 & - & - \\
    \citet{lee2021direct} (trained w/ text)           &    - &               - & - &  39.9 & 3.41 $\pm$ 0.14 & - \\
    \bottomrule
\end{tabular}
\label{tbl:performance}
\end{small}
\end{table*}

To evaluate the translation quality, we used the same two datasets as in \citet{jia2019direct}, both of which have translation speech in a single female speaker's voice.
Following \citet{jia2019direct}, the translation quality is measured by BLEU on ASR transcription from the translation speech (in lowercase, excluding punctuation marks except for apostrophes), compared to reference translation text.
Because ASR makes errors, such BLEU can be thought of as a lower bound of the translation quality.
We used an ASR model from \citet{park2020improved}, trained on LibriSpeech \citep{panayotov2015librispeech} and LibriLight \citep{kahn2020libri} corpora.
For a fair comparison, we retrained the baseline Translatotron models and evaluated them using the same ASR model.
The same ST$\to$TTS cascade S2ST baselines from \citet{jia2019direct} were used and re-evaluated, which were composed of strong ST models and a Tacotron 2 TTS model.
The predicted mel-spectrogram is converted to waveform using the Griffin-Lim algorithm for all models.

As shown in Table~\ref{tbl:performance}, the translation quality from Translatotron~2 outperformed the original Translatotron by $+$15.5 BLEU on Fisher Es-En and $+$5.2 BLEU on Conversational. Applying ConcatAug further improved the performance on the smaller Fisher Es-En dataset by $+$0.5 BLEU. These improvements narrowed down the performance gap between end-to-end direct S2ST and cascade S2ST from 16.4 / 8.4 down to 0.4 / 3.7 BLEU on the two datasets respectively.

\subsection{Speech naturalness}
\label{sec:exp-naturalness}

The naturalness of the predicted translation speech is evaluated by subjective listening test, reporting 5-scale mean opinion scores (MOS) with 95\% confidence interval on 1,000 randomly sampled predictions. A WaveRNN-based neural vocoder was used for converting the mel-spectrograms predicted from S2ST models to waveforms.

As shown in Table~\ref{tbl:performance}, the naturalness of the translation speech predicted from Translatotron 2 is significantly better than from the original Translatotron, and is on-par with or very close to the cascade systems which used one of the state-of-the-art TTS models, Tacotron 2, for synthesizing translation speech from text.

Consistent with \citet{jia2019direct}, despite that the training targets in the Conversational dataset is synthesized with a lower quality Griffin-Lim vocoder, the trained S2ST model is able to produce translation speech in significantly higher naturalness  when a higher quality neural vocoder is used at inference time.

\subsection{Speech robustness}

We specifically evaluated the robustness issue of over-generation in the predicted translation speech, such as babbling or long pause, measured by unaligned duration ratio (UDR) \citep{shen2020non} with a 1-second threshold.\footnote{Under-generation
\ifdefined\isaccepted
(i.e. WDR from \citet{shen2020non})
\fi
does not apply because of the nature of translation. Related errors are reflected in the BLEU evaluation.}
The ASR transcription from the translation speech is used for alignment, using a confidence islands-based forced alignment model \citep{chiu2017speech}.

As can be seen from Table~\ref{tbl:performance},
the UDR from Translatotron~2 is about 7 and 4 times lower than from Translatotron on the Fisher Es-En and Conversational datasets, respectively. It is even about 3 times lower than the training targets from the Conversational set, while is about the same as the training targets from Fisher Es-En. This can be explained by the fact that the training targets in the Conversational set were synthesized by the Tacotron 2 TTS model, which by itself suffered from over-generation, while the same in Fisher Es-En were synthesized by a more robust Parallel WaveNet \citep{oord2018parallel} TTS model (see Table~\ref{tbl:datasets}).
The results suggest that Translatotron 2 drastically improves robustness than Translatotron, and is also robust to a small ratio of disfluency in the training targets.

\subsection{Voice preservation}
\label{sec:exp-voice-retention}

\begin{table}[t]
\centering
\begin{small}
\caption{S2ST performance with voice preservation on Conversational dataset. Speaker similarity MOS is evaluated between Spanish source speech and English translation speech. (Numbers not directly comparable to Table~\ref{tbl:performance} because of dataset differences.)}
\begin{tabular}{lccc}
    \toprule
     & BLEU & Naturalness & Similarity  \\
    \midrule
    \multicolumn{4}{l}{\kern-0.5em\emph{Proposed:}}\\
    Translatotron 2                  & \B 57.3 & \B 3.24 $\pm$ 0.08 & \B 2.33 $\pm$ 0.08 \\
    \quad + ConcatAug                & 56.8 & 2.94 $\pm$ 0.08 & 2.12 $\pm$ 0.07 \\
    Translatotron                    & 48.5 & 2.55 $\pm$ 0.09 & 2.30 $\pm$ 0.07 \\
    \quad + ConcatAug                & 51.3 & 2.76 $\pm$ 0.09 & 2.19 $\pm$ 0.07 \\
    \cmidrule(lr){1-4}
    Reference (synthetic)                 & 81.3 & 3.40 $\pm$ 0.08 & 2.55 $\pm$ 0.07 \\
    \midrule
    \multicolumn{4}{l}{\kern-0.5em\emph{\citet{jia2019direct}:}}\\
    Translatotron     & 36.2 & $ 3.15 \pm 0.08$ & $ 1.85 \pm 0.06$ \\
    Reference (human)  & 59.9 & $4.10 \pm 0.06$ & - \\
    \bottomrule
\end{tabular}
\label{tbl:voice-transfer}
\end{small}
\end{table}

\begin{table*}[t]
\centering
\begin{small}
\caption{Voice preservation performance on speaker turns. Speaker similarity MOS between the leading/trailing 1.6-second segment from the English translation speech and the entire 1st/2nd source speaker's Spanish speech is reported. ($\uparrow$ / $\downarrow$: higher/lower values are better.)}
\begin{tabular}{lccccc}
    \toprule
     \multirow{1}{*}{}& \multicolumn{2}{c}{1st source speaker} & \multicolumn{2}{c}{2nd source speaker} \\
    \cmidrule(r){2-3}\cmidrule(l){4-5}
    &                 Leading seg. $\uparrow$   & Trailing seg. $\downarrow$  & Leading seg. $\downarrow$ & Trailing seg. $\uparrow$ \\
    \midrule
    Translatotron 2      & 2.22 $\pm$ 0.07 & 2.15 $\pm$ 0.07 & 2.04 $\pm$ 0.07 & 2.00 $\pm$ 0.07 \\
    \quad + ConcatAug    & \B 2.44 $\pm$ 0.07 & 1.82 $\pm$ 0.07 & \B 1.76 $\pm$ 0.07 & \B 2.51 $\pm$ 0.08 \\
    Translatotron        & 1.87 $\pm$ 0.06 & 1.90 $\pm$ 0.07 & 2.06 $\pm$ 0.07 & 2.05 $\pm$ 0.07 \\
    \quad + ConcatAug    & 2.18 $\pm$ 0.07 & \B 1.71 $\pm$ 0.06 & 1.93 $\pm$ 0.07 & 2.35 $\pm$ 0.07 \\
    \midrule
    Reference (synthetic)     & 2.58 $\pm$ 0.08 & 1.62 $\pm$ 0.06 & 1.83 $\pm$ 0.07 & 2.44 $\pm$ 0.07 \\
    \bottomrule
\end{tabular}
\label{tbl:speaker-turns}
\end{small}
\end{table*}

To evaluate the ability of preserving speakers' voices while translating their speeches from one language to another, we augmented the Conversational dataset by synthesizing target speech using a voice-transferring TTS model as described in Sec.~\ref{sec:multispeaker_data}. Examples with source speech shorter than 1 second were discarded for the stability of voice transferring. The result dataset contains parallel utterances with similar voices on both sides.
S2ST models were trained on this dataset without any explicit conditioning on speaker embeddings or IDs (i.e. no speaker encoder for the original Translatotron).
Following \citet{jia2019direct}, we reduced the pre-net dimension of the synthesizer to 16 to encourage it to infer voice information from the encoder output instead of from the teacher-forcing inputs.

5-point subjective MOS on both naturalness and speaker similarity was evaluated with 1,000 random samples or pairs of samples from the test set,
reported with 95\% confidence interval. As Table~\ref{tbl:voice-transfer} shows,
when the proposed approach for voice preservation was used,
both Translatotron 2 and Translatotron obtained about the same speaker similarity MOS as the original Translatotron but significantly better translation quality. Translatotron 2 further outperformed Translatotron in terms of translation quality and speech naturalness, which is consistent with the experimental results for translating in a single speaker's voice (Sec.~\ref{sec:exp-translation-quality}, \ref{sec:exp-naturalness}). 
It is worth to note that
the speaker similarity from S2ST models is capped by the same of the training targets, which by itself is limited.
This can be partially due to the performance of the voice-transferring TTS model used for synthesizing the training targets, and partially due to the fact that cross-lingual speaker similarity evaluation is more challenging to raters (some rating comments are purely based on language difference), as also observed in \citet{zhang2019learning}. 
Obtaining better quality training targets, such as human recordings instead of synthesized speech, may further improve the performance of voice preservation with the proposed approach.

\subsubsection{Speaker turns}
\label{sec:exp-speaker-switching}

Speaker similarity evaluation with speaker turns on entire translation speech is challenging because it would require speaker separation on both source and target speeches. The content re-ordering during translation and translation errors would also add extra difficulty. We approximated by considering the leading/trailing short segments in the translation speech as corresponding to each of the two speakers in the source speech with a single speaker turn.

We trained Translatotron 2 and Translatotron on the dataset described in Sec.~\ref{sec:exp-voice-retention}, with half of the training examples augmented by ConcatAug.
The evaluation set was artificially constructed in a similar way by applying ConcatAug, so that each utterance contains two speakers' voices. 
We evaluated subjective speaker similarity MOS  between the two entire source utterances before ConcatAug and the leading/trailing 1.6-second segments
from the translation speech.
Evaluation examples with target speech shorter than 2~seconds before ConcatAug were discarded.

As can be seen from Table~\ref{tbl:speaker-turns}, the impact of ConcatAug is consistent on Translatotron 2 and Translatotron. When ConcatAug was not used during training,
for each source speaker, the similarity compared to the leading/trailing segment from the translation speech was about the same; and for each segment in the translation speech, the speaker similarity compared to the first/second source speaker was also close. This suggests that the translation speech imitated both source speakers at the same time regardless of the speaker turn.
When ConcatAug was used, both models obtained significantly higher speaker similarity on matched pairs than mismatched pairs, indicating that the models successfully separated two speakers and preserved voices for each of them respectively.
It can also be seen that Translatotron~2 obtained significantly higher speaker similarity than Translatotron on matched pairs, indicating the effectiveness of Translatotron 2.

Such quantitative evaluation cannot reflect how the predicted translation speech transits from one speaker's voice to another speaker's. Listening to audio samples 
\ifdefined\isaccepted
(available online)
\else
(available in the supplemental materials)
\fi
verified that the voice changed instantly on sentence boundaries without blurry, rather than a smoothed transition. %
A sample of S2ST on such a speaker turn from Translatotron~2 is visualized in Figure~\ref{fig:speaker-switching}.

While ConcatAug enables S2ST models to preserve speakers' voices on speaker turns and improves translation quality on small datasets, it may negatively impact the speech naturalness and speaker similarity on models with strong performance, as shown in Table~\ref{tbl:performance} and Table~\ref{tbl:voice-transfer}. It may be explained by the fact that the augmented utterances sound less natural and may involve abrupt change in volume and background noise on the artificial speaker turns. This suggests headroom for improvement.

\subsection{Multilingual S2ST}
\label{sec:exp-multilingual}

We also conducted experiments to evaluate the performance of multilingual X$\to$En S2ST. We trained Translatotron~2 and Translatotron on the 4 high-resource language pairs from the CoVoST 2 corpus \citep{wang2020covost}, using TTS synthesized target speech in a single female speaker's voice.\footnote{An expanded version of this dataset is released as the CVSS corpus \citep{jia2022cvss}. However, the results are not directly comparable because of different data splits and reference texts.%
} The original Common Voice \citep{ardila2020common} data split instead of the CoVoST 2 data split was followed. The models were not explicitly conditioned on languages.
For a fair comparison, both models used SpecAugment, but did not use auxiliary supervision from the source phonemes.

The translation quality as measured by BLEU on ASR transcription from the translation speech is shown in the first rows of each block in Table~\ref{tbl:multilingual}.  Translatotron 2  outperformed Translatotron by $+$9.4 BLEU on average on the 4 language pairs. Although the BLEU scores are not directly comparable between S2ST and ST (because of ASR transcription and BLEU calculation difference), the close numbers suggest that Translatotron~2 obtained translation quality comparable to the baseline ST model.

\subsubsection{Ablation studies}
\label{s:ablation}

\begin{table}[t]
\centering
\begin{small}
\caption{Ablation studies of multilingual X$\to$En S2ST on 4 high-resource language pairs from CoVoST 2, measured by BLEU on ASR transcription from the translation speech. $+$ / $-$ indicates using or replacing a component (see Sec.~\ref{s:ablation}).}
\setlength{\tabcolsep}{0.6em}
\begin{tabular}{lrrrr}
    \toprule
      & fr & de & es & ca \\
    \midrule
    Translatotron 2                & \B 27.0 & \B 18.8 & \B 27.7 & \B 22.5 \\
    \quad $-$ SpecAugment          & 25.9 & 17.9 & 25.9 & 21.8 \\
    \quad $-$ Conformer encoder    & 26.4 & 18.1 & 26.4 & 21.8 \\
    \quad $-$ NAT synthesizer      & 26.9 & 18.3 & 27.0 & 22.0 \\

    \midrule
    Translatotron (w/ SpecAugment) & 17.7 & 9.9 & 17.7 & 13.1 \\
    \quad $+$ Conformer encoder    & 18.9 & 10.8 & 18.8 & 13.9  \\
    \quad\quad $+$ NAT synthesizer     &  4.0 &  2.1 &  3.5 &  2.5 \\
    \midrule
    ST \citep{wang2020covost} & 27.0 & 18.9 & 28.0 & 23.9 \\
    \midrule
    Reference (synthetic) & 82.1 & 86.0 & 85.1 & 89.3 \\
    \bottomrule
\end{tabular}
\label{tbl:multilingual}
\end{small}
\end{table}

To understand the importance of each component in Translatotron 2, we conducted ablation studies on this multilingual X$\to$En dataset. All models in the ablation used the same input and output features, SpecAugment settings, and learning rate schedules (detailed in Appendix \ref{a:hparams}). No auxiliary supervision from source text was used.
For models not using a Conformer encoder, we first applied the same 4$\times$ temporal subsampling as in the Conformer encoder, then used a 256$\times$8 bidirectional LSTM stack to encoder the subsampled features. The number of parameters in this LSTM encoder is close to the same in the Conformer encoder.
For the Translatotron model using a NAT synthesizer,
the same hyperparameters as in Translatotron 2 were used.
For Translatotron 2 not using a NAT synthesizer, a non-autoregressive Conformer synthesizer (Sec.~\ref{s:exp-nar-synthesizer}) was used.
All the rest hyperparameters followed Appendix~\ref{a:hparams} for Translatotron 2, and followed the Conversational model from \citet{jia2019direct} for Translatotron. All models were trained for 200K steps with a batch size of 768. The checkpoints for evaluation were picked by the best average BLEU on 4 language pairs on the validation set.

The results are shown in Table~\ref{tbl:multilingual}. As can be seen, while the use of Conformer, SpecAugment, and NAT synthesizer helps the performance of Translatotron 2, replacing them with alternative architectural choices or removing SpecAugment only reduced the performance by a small degree ($<$2~BLEU). Similarly, directly using these components in Translatotron does not bring its performance close to Translatotron~2. These results suggest that the improvements of Translatotron~2 primarily comes from the high-level architectural design which addressed the performance bottlenecks existing in Translatotron (Sec.~\ref{sec:method}), rather than the choices of each individual component.

\subsection{Non-autoregressive synthesizer}
\label{s:exp-nar-synthesizer}

\begin{table}[t]
\centering
\begin{small}
\caption{Ablation studies on Conversational dataset using an autoregressive RNN + Conv synthesizer and a non-autoregressive Conformer synthesizer.}
\begin{tabular}{lcc}
    \toprule
    Synthesizer & BLEU & Naturalness \\
    \midrule
    RNN + Conv       & 55.6 & 4.21 $\pm$ 0.06 \\
    Conformer  & 54.5 & 3.61 $\pm$ 0.09 \\
    \bottomrule
\end{tabular}
\label{tbl:non-ar-synthesizer}
\end{small}
\end{table}

It is tempting to use a non-autoregressive architecture for the synthesizer of Translatotron~2, which may significantly reduce its inference latency, similar to recent works on non-autoregressive TTS \citep{ren2019fastspeech,ren2020fastspeech2,guo2021recent,lee2021bidirectional,elias2020parallel,elias2021parallel}.
We experimented with using a 6-layer Conformer synthesizer \citep{guo2021recent} with a dimension of 512 and 8 attention heads on both Conversational and CoVoST 2 datasets. 

As can be seen from Table~\ref{tbl:multilingual} and \ref{tbl:non-ar-synthesizer}, using a Conformer-based non-autoregressive synthesizer obtained comparable translation quality to using an autoregressive NAT synthesizer (with BLEU on ASR transcription up to 1.1 BLEU lower). However, it caused a significant regression on the naturalness of the predicted translation speech, which is consistent with the observation in TTS in \citet{shen2020non,peng2020non,hwang2021tts}, etc., suggesting more exploration is needed on this direction.

%% file: 4-conclusion.tex
\section{Conclusion}

We proposed \emph{Translatotron 2}, a neural direct S2ST model that can be trained end-to-end.
Experimental results on three datasets consistently suggest that Translatotron~2 outperforms the original Translatotron by a large margin on both translation quality (up to $+$15.5 BLEU) and speech generation quality, and approaches cascade S2ST.%

In addition, we proposed a simple method for preserving speakers' voices from the source speech to the translation speech in a different language. Unlike existing approaches, the proposed method is able to preserve each speaker's voice on speaker turns without requiring for speaker segmentation.
Furthermore, compared to existing approaches, it better preserves speaker's privacy and mitigates potential misuse of voice cloning for creating spoofing audio artifacts.

Future works include extending Translatotron 2 to support simultaneous translation, cross-lingual prosody transfer, unwritten languages, and further quality improvement by utilizing self-supervised pre-training \citep{baevski2020wav2vec,wang2021large} and weakly supervised data~\citep{jia2019leveraging}.

\ifdefined\isaccepted 
\section*{Acknowledgements}
The authors would like to thank Chung-Cheng Chiu, Quan Wang, Heiga Zen, Ron J. Weiss, Wolfgang Macherey, Yu Zhang, Yonghui Wu, Hadar Shemtov, Ruoming Pang, Nadav Bar, Michael Hassid, and the rest of the Google Research team for helpful discussions and previous work on data preparation.
\fi

%% file: 9-appendix.tex
\appendix

\onecolumn

\section{Table of hyper-parameters}
\label{a:hparams}

\begin{table*}[h!]
\centering
\begin{small}
\caption{Model hyper-parameters used in the experiments. (``$\times n$'': $n$ layers; $^\dagger$: 128-dim pre-net is used for translating in a single voice; 16-dim pre-net is used for voice preservation.)}
\label{tbl:model-params}
\begin{tabular}{lccc}
\toprule
     & Fisher Es-En & CoVoST 2 & Conversational \\
    \midrule
    \emph{Input} \\
    Sample rate (Hz) & 8,000 & 48,000 & 16,000 -- 48,000 \\
    Mel channels        & \multicolumn{3}{c}{80} \\
    Mel lower band (Hz) & \multicolumn{3}{c}{125} \\
    Mel upper band (Hz) & 3,800 & 7,600 & 7,600 \\
    Frame size (ms)     & \multicolumn{3}{c}{25.0} \\
    Frame step (ms)     & \multicolumn{3}{c}{10.0} \\
    \midrule
    \emph{Output} \\
    Sample rate (Hz)    & \multicolumn{3}{c}{24,000} \\
    Mel channels        & \multicolumn{3}{c}{128} \\
    Mel lower band (Hz) & \multicolumn{3}{c}{20} \\
    Mel upper band (Hz) & \multicolumn{3}{c}{12,000} \\
    Frame size (ms)     & \multicolumn{3}{c}{50.0} \\
    Frame step (ms)     & \multicolumn{3}{c}{12.5} \\
    \midrule
    \emph{SpecAugment} \\
    Freq blocks & \multicolumn{3}{c}{2} \\
    Time blocks & \multicolumn{3}{c}{10} \\
    Freq block max length ratio & \multicolumn{3}{c}{0.33} \\
    Time block max length ratio & \multicolumn{3}{c}{0.05} \\
    \midrule
    \emph{Encoder} \\
    Conformer dims    & \multicolumn{3}{c}{144 $\times$ 16} \\
    Attention heads & \multicolumn{3}{c}{4} \\
    Conv kernal size & \multicolumn{3}{c}{32} \\
    Subsample factor & \multicolumn{3}{c}{4} \\
    \midrule
    \emph{Attention} \\
    Output dim & 256 & 512 & 512 \\
    Hidden dim & 512 & 512 & 512 \\
    Attention heads & 4 & 8 & 8 \\
    Dropout prob & 0.1 & 0.2 & 0.2 \\
    \midrule
    \emph{Decoder} \\
    LSTM dims & 256 $\times$ 4 & 512 $\times$ 6 & 512 $\times$ 4 \\
    Zoneout prob & 0.1 & 0.1 & 0.1 \\
    Phoneme embedding dim & 96 & 256 & 256 \\
    Label smoothing uncertainty & 0.1 & 0.1 & 0.1 \\
    Loss weight & 10.0 & 10.0 & 10.0 \\
    \midrule
    \emph{Duration predictor} \\
    Bi-LSTM (dim $\times$ layers) & 64 $\times$ 2 & 128 $\times$ 2 & 128 $\times$ 2 \\
    Loss weight & 1.0 & 1.0 & 1.0 \\
    \midrule
    \emph{Synthesizer} \\
    LSTM dims  & \multicolumn{3}{c}{1,024 $\times$ 2} \\
    LSTM zoneout prob & \multicolumn{3}{c}{0.1} \\
    Pre-net dims & 128 $\times$ 2 & 128 $\times$ 2 & 128 / 16 $^\dagger$ $\times$ 2  \\
    Pre-net dropout prob & \multicolumn{3}{c}{0.5} \\
    Post-net (kernel, channels) $\times$ layers &  \multicolumn{3}{c}{(5, 512) $\times$ 4  $+$ (5, 128)} \\
    Loss weight & \multicolumn{3}{c}{1.0} \\
    \midrule
    \multicolumn{4}{l}{\emph{Training}} \\
    Optimizer & \multicolumn{3}{c}{Adam \citep{kingma2015adam}}  \\
    Learning rate schedule & \multicolumn{3}{c}{\citet{vaswani2017attention}} \\
    Learning rate (peak) & 4.2$\times10^{-3}$ & 2.2$\times10^{-3}$ & 3.3$\times10^{-3}$ \\
    Warm-up steps & 10K & 20K & 10K \\
    Batch size & 1,024 & 768 & 768 \\
    $L^2$ regularization weight & $10^{-6}$ & $10^{-6}$ & $10^{-6}$ \\
    \bottomrule
\end{tabular}
\end{small}
\end{table*}

\section{Objective speaker similarity analysis}
\label{a:similarity}

\begin{figure*}[t]
    \centering
    \setlength{\tabcolsep}{0ex}
    \begin{tabular}{ccccc}
        \includegraphics[width=0.2\linewidth]{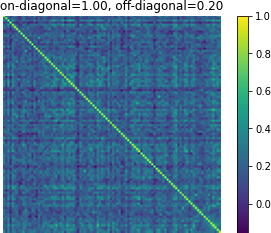} &
        \includegraphics[width=0.2\linewidth]{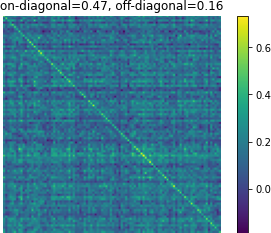} &
        \includegraphics[width=0.2\linewidth]{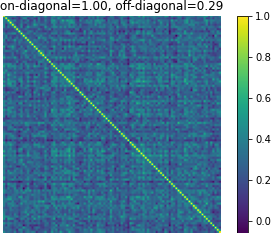} &
        \includegraphics[width=0.2\linewidth]{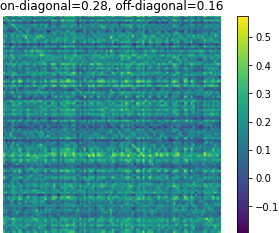} & 
        \includegraphics[width=0.2\linewidth]{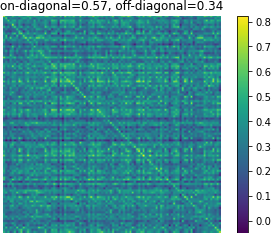} \\
        $src$ vs $src$\quad~ &
        $tgt$ vs $src$\quad~ &
        $tgt$ vs $tgt$\quad~ &
        $s2st$ vs $src$\quad~ & 
        $s2st$ vs $tgt$\quad~ \\
    \end{tabular}
    \caption{Affinity matrices of d-vector similarity among 100 random examples. (``s2st'' refers to the predictions from Translatotron 2.)}
    \label{fig:affinity-mat}
\end{figure*}

\begin{table*}[t]
\centering
\begin{small}
\caption{Objective d-vector similarity between the predicted translated speech (English) and the source human speech (Spanish) on speaker turns. The similarity between the leading/trailing 1.6-second segment from the predicted speech and the entire 1st/2nd source speaker's speech is measured. ($\uparrow$ / $\downarrow$ means higher/lower values are better.)}
\label{tbl:speaker-turns-dvector}
\begin{tabular}{lccccc}
    \toprule
     \multirow{1}{*}{}& \multicolumn{2}{c}{1st source speaker} & \multicolumn{2}{c}{2nd source speaker} \\
    \cmidrule(r){2-3}\cmidrule(l){4-5}
    &                 Leading seg. $\uparrow$   & Trailing seg. $\downarrow$  & Leading seg. $\downarrow$ & Trailing seg. $\uparrow$ \\
    \midrule
    Translatotron 2         %
    & 0.21             & 0.19           & 0.21            & 0.19 \\
    \quad + ConcatAug     %
    & 0.20             & \B 0.14  & \B 0.14   & 0.21 \\
    Translatotron           %
    & 0.20             & 0.22           & 0.27            & 0.29 \\
    \quad + ConcatAug     %
    & \B 0.32    & 0.16           & \B 0.14   & \B 0.35 \\
    \midrule
    Reference (synthetic)      & 0.48 & 0.17 & 0.15 & 0.48 \\
    \bottomrule
\end{tabular}
\end{small}
\end{table*}

Subjective speaker similarity evaluation is costly and has a long turnaround. We explored alternative objective evaluation using separately trained speaker encoders, such as d-vector \citep{wan2018generalized}. We evaluated the voice retention performance using the cosine similarity of the d-vectors.

We first checked the scenario that each input contains a single speaker's recording. Figure~\ref{fig:affinity-mat} visualizes the affinity matrices of d-vector similarity among different input utterances for a Translatotron 2 model. The outstanding higher similarity values on the diagonals indicate that the model is able to preserve the source speaker's voice in the predicted translation speech.

We then conducted a detailed evaluation for the voice retention performance for speaker turns. The experiment setting up was identical to Section~\ref{sec:exp-speaker-switching}, except that the speaker similarity was measured by d-vector similarity instead of subjective MOS evaluation.
The d-vectors for each source speaker were computed on the entire original utterance before concatenation; the d-vectors for each speaker in the prediction is approximated by computing on the leading/trailing 1.6 seconds of predicted speech.

The results are shown in Table~\ref{tbl:speaker-turns-dvector}. Consistent with the MOS evaluation results in Table~\ref{tbl:speaker-turns}, when the concatenation augmentation was not used, the d-vector similarity to each source speaker is about the same regardless if it was compared to the leading or trailing segments, 
indicating that the predicted speech was in a single speaker's voice and the model was unable to separate different speakers in the input, but rather optimized for both source speakers at the same time.
When the concatenation augmentation was used, the d-vector similarity was significantly higher between matched pairs than between unmatched pairs, indicating that the models were able to separate different speakers in the input and preserve their voices in the predicted translation speech respectively.

However, when these similarities are compared among different models, it seems to suggest that Translatotron performed better than Translatotron 2, which is contradictory to the subjective evaluation results in Table~\ref{tbl:speaker-turns}. By carefully listening to the audio samples, we found that such discrepancy may be due to the fact that the d-vector model was also sensitive to non-voice related acoustic characteristics, such as reverb and channel noise in the audios. This is likely a consequence of the fact that in the large-scale training set for the d-vector model used in the evaluation, each speaker is typically associated with a particular recording condition, e.g. recording device and room.
Because the encoder output from the Translatotron model was of significantly larger dimension than from the Translatotron 2 model (2048 vs 144), it was capable of carrying more non-voice acoustic information and thus obtained better d-vector similarity, which not necessarily indicating higher speaker similarity.

These results suggest that while such speaker encoder-based objective analysis reveals insightful indications about the performance of the S2ST models, it can be less reliable compared to subjective MOS evaluation. Such reliability also highly depends on the training details of the speaker encoder model being used, especially the training corpus.